\documentclass{article}

\usepackage{PRIMEarxiv}

\usepackage[utf8]{inputenc} 
\usepackage[T1]{fontenc}    
\usepackage{hyperref}       
\usepackage{url}            
\usepackage{booktabs}       
\usepackage{amsfonts}       
\usepackage{nicefrac}       
\usepackage{microtype}      
\usepackage{lipsum}
\usepackage{fancyhdr}       
\usepackage{graphicx}       
\usepackage{subfig}
\usepackage{orcidlink}
\usepackage{graphicx}
\usepackage{booktabs}
\usepackage{tabularray}
\UseTblrLibrary{diagbox}
\usepackage{colortbl} 
\usepackage[ruled,vlined]{algorithm2e}
\usepackage{algpseudocode}
\usepackage{wrapfig}
\usepackage[ruled,vlined]{algorithm2e}
\usepackage{graphicx}
\usepackage{textcomp}
\usepackage{fancyhdr,graphicx,amsmath,amssymb}
\usepackage{algpseudocode}
\usepackage{multirow}
\usepackage{graphicx}
\usepackage{cleveref}

\graphicspath{{media/}}     

\title{{GSTAM}: Efficient Graph Distillation with Structural Attention-Matching
\thanks{To be presented at ECCV-DD 2024} 
}

\author{
  Arash Rasti-Meymandi, Ahmad Sajedi \\
  University of Toronto, ECE Dept. \\
  Toronto,Canada\\
  \texttt{\{arash.rasti,ahmad.sajedi\}@mail.utoronto.com} \\
   \And
  Zhaopan Xu \\
National University of Singapore
  Singapore\\
  \texttt{e0983526@u.nus.edu} \\
   \And
   Konstantinos N. Plataniotis\\
  University of Toronto, ECE Dept. \\
  Toronto,Canada\\
  \texttt{kostas@ece.utoronto.ca} \\
}

\begin{document}
\maketitle

\begin{abstract}
Graph distillation has emerged as a solution for reducing large graph datasets to smaller, more manageable, and informative ones. Existing methods primarily target node classification, involve computationally intensive processes, and fail to capture the true distribution of the full graph dataset. To address these issues, we introduce Graph Distillation with Structural Attention Matching (\texttt{GSTAM}), a novel method for condensing graph classification datasets. \texttt{GSTAM} leverages the attention maps of GNNs to distill structural information from the original dataset into synthetic graphs. The structural attention-matching mechanism exploits the areas of the input graph that GNNs prioritize for classification, effectively distilling such information into the synthetic graphs and improving overall distillation performance. Comprehensive experiments demonstrate \texttt{GSTAM}'s superiority over existing methods, achieving $0.45\%$ to $6.5\%$ better performance in extreme condensation ratios, highlighting its potential use in advancing distillation for graph classification tasks (Code available at \href{https://github.com/arashrasti96/GSTAM}{github.com/arashrasti96/GSTAM}).
\end{abstract}


\section{Introduction} 
The focus on graph-structured information has increased dramatically over the past decade due to the rise of social media, recommendation systems, and various real-world structures like molecules, 3D meshes, and brain connectivity. Today, graph-structured datasets, such as knowledge graphs \cite{hu2020open, ching2015one} and e-commerce platforms \cite{ying2018graph}, have grown to encompass millions of nodes and billions of edges. This growth introduces significant challenges for both training Graph Neural Networks (GNNs) and exploring new research areas, such as continual learning \cite{zhang2022cglb}, neural architecture search \cite{gao2021graph, zhang2023dynamic, xu2023not}, and knowledge amalgamation \cite{jing2021amalgamating}.
 
To address these issues, one effective solution is graph condensation (also known as graph distillation), which involves reducing the size of a large graph dataset to a smaller, more manageable one. Graph condensation can be categorized into four primary optimization approaches. (i) Coreset selection involves selecting a representative subset of graphs from the original dataset, although it typically does not consider the specific requirements of downstream tasks. (ii) Gradient matching, as explored in \cite{jin2022condensing, jin2021graph}, focuses on aligning the gradients of network parameters between synthetic and real graph data. Despite their success, these methods rely on a bi-level optimization problem with inner and outer loops, which are challenging to solve. (iii) Trajectory matching, discussed in \cite{zheng2024structure, zhang2024navigating}, aligns the training trajectories of a student GNN on the condensed graph with those of multiple teacher GNNs on the original graph. However, trajectory matching is time- and memory-intensive, requiring training and using multiple expert models during optimization. (iv) Distribution matching, introduced in \cite{liu2022graph}, creates a smaller graph that preserves the distribution of receptive fields of the original graph, using a loss measured by maximum mean discrepancy (MMD). Although this method reduces computation costs, it tends to underperform compared to methods such as trajectory matching. 

Most methods mentioned above focus primarily on the node classification task, aiming to condense ``one'' large graph into a smaller graph. However, another significant category of graph problems is graph classification. In this case, it is necessary not only to reduce the number of graph samples but also to reduce the number of nodes within each sample. This requirement makes approaches like distribution matching, as described in \cite{liu2022graph}, less feasible for graph classification scenarios. Additionally, most existing distillation works on graph classification datasets ignore the significance of cross-architecture analysis, which indicates the generalizability of the learned synthetic graphs. Furthermore, these approaches fail to capture the network's attention on the input graph in different layers, which, as shown in \cite{zagoruyko2016paying}, contains rich information useful for distillation purposes. 

\begin{figure}[tb!]
    \centering
    \subfloat[]{
        \includegraphics[width=.9\textwidth]{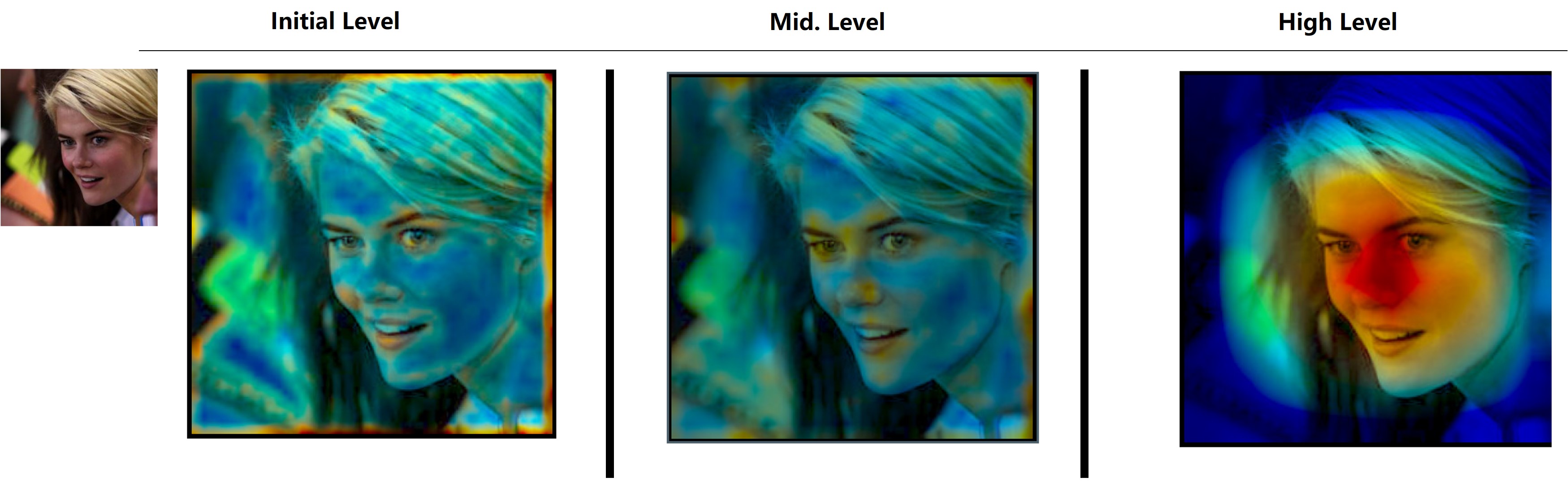}
        \label{fig:attention2}
    }
    \hfill
    \subfloat[]{
        \includegraphics[width=1\textwidth]{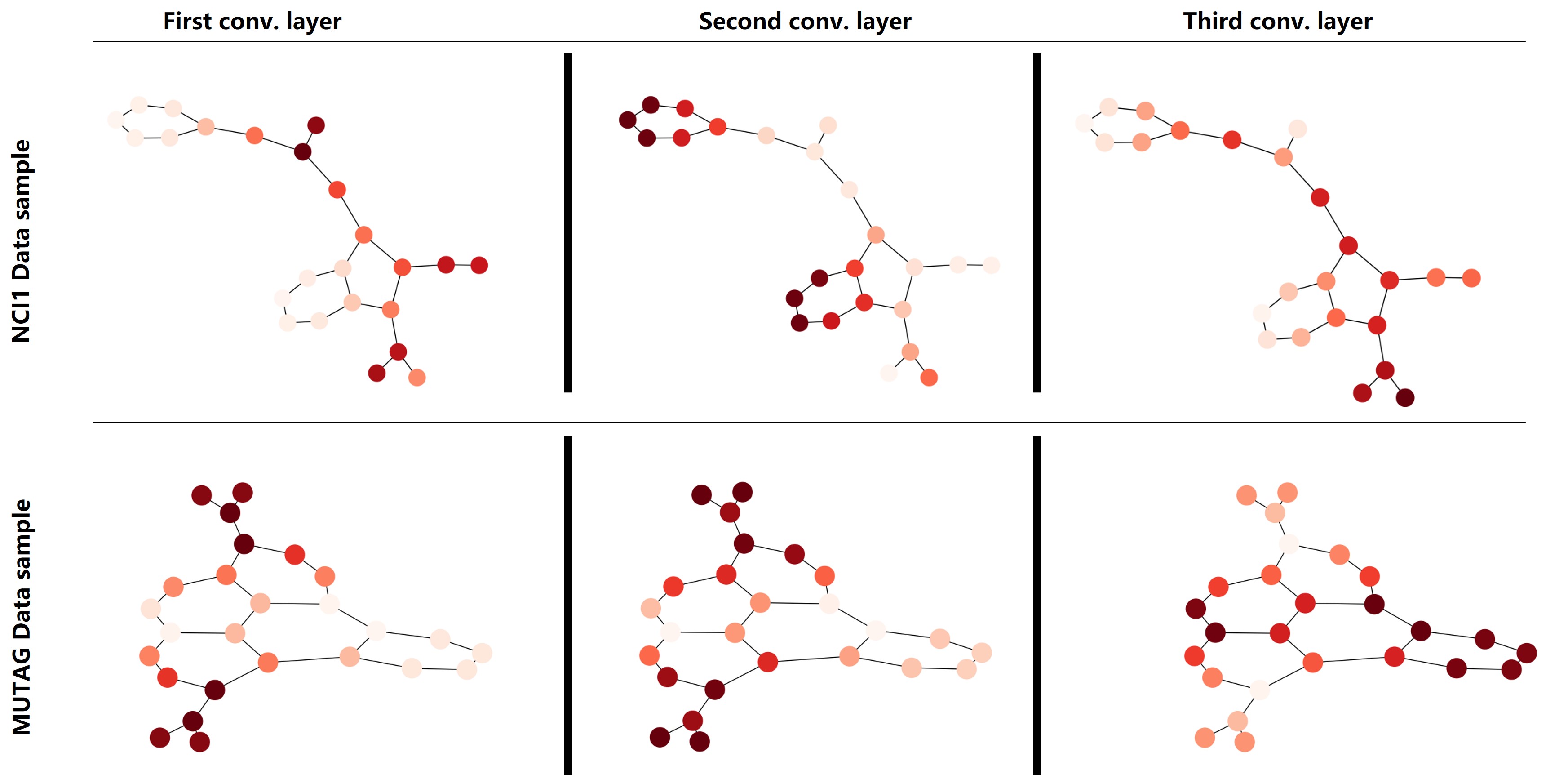}
        \label{fig:attention}
}
    \caption{\textbf{Motivating Example:} (a) The attention maps over different levels of a network trained for face recognition that indicates the focus of the network on the particular input image \cite{zagoruyko2016paying}. The brighter the color, the greater the network's focus on that specific part of the image. (b) An input graph and its corresponding structural attention map created using a technique similar to Grad-CAM \cite{selvaraju2017grad} can reveal where different layers of a trained GNN focus to classify the given graph. This information is valuable when distilling a graph dataset, as it highlights the areas of the input graph that the GNN prioritizes for classification. A darker red color represents higher attention. }
    \label{fig:overall_attention}
\end{figure}

It is shown in the literature \cite{zagoruyko2016paying, sajedi2023datadam, khaki2024atom} that the different layers of CNN prioritize different parts of the input image which are useful information for the classification task (see \cref{fig:overall_attention} (a)). Motivated by such findings, we extend the attention mechanism to the Graph classification task (see \cref{fig:overall_attention} (b)). To that aim, we introduce \textbf{G}raph Distillation with \textbf{ST}ructural \textbf{A}ttention \textbf{M}atching (\texttt{GSTAM}), a method that distills the attention maps of GNNs trained on full graph datasets into synthetic graphs to extract rich information from the full dataset. Our approach specifically targets the graph classification task. Unlike previous methods such as \cite{jin2022condensing}, \texttt{GSTAM} does not rely on gradient-based techniques and is more time-efficient compared to trajectory matching algorithms. Moreover, \texttt{GSTAM} aims to find a better representation of the input graphs' distribution for the distillation process. The key contributions of the proposed methodology are as follows:

\textbf{[C1]} We introduce a novel Structural Attention Matching mechanism (STAM) designed for GNNs, which addresses the variability of graph samples to enable effective distillation. To the best of our knowledge, this is the first use of attention maps in GNNs for graph distillation purposes.

\textbf{[C2]} We empirically demonstrated that no bi-level optimization is required to distill graph datasets by introducing a new loss function. This function minimizes the attention maps of multiple randomly initialized parameters of GNNs, fed by both the full and synthetic graph datasets, without necessitating the learning of parameters.

\textbf{[C3]} We conduct extensive experiments on various graph datasets as well as cross-architecture analysis focusing on graph classification problems, comparing our method against coreset-based approaches and state-of-the-art distillation algorithms. Our proposed method consistently demonstrates superior performance in most cases.

\subsection{Related Works}
\textbf{Coreset selection.} Coreset selection is an early data-centric technique designed to efficiently identify a representative subset from a full dataset, enhancing the performance and efficiency of downstream training. Various approaches have been developed over time, including geometry-based methods \cite{agarwal2020contextual,sener2018active}, loss-based techniques, decision-boundary-focused strategies, bilevel optimization methods, and gradient-matching algorithms. Among these approaches are several notable ones: Random: Selects samples randomly to form the coreset. Herding\cite{welling2009herding}: Chooses samples closest to the cluster center. K-Center\cite{farahani2009facility,sener2018active}: Select multiple center points to minimize the maximum distance between data points and their nearest center. While these methods have demonstrated moderate success in improving training efficiency, they face inherent limitations in capturing comprehensive information. Treating each image in the selected subset independently results in a lack of rich features that could be obtained by considering the diversity within classes. These limitations have driven the development of dataset distillation in the field.

\textbf{Graph distillation.}  Unlike traditional subset selection methods, the idea of graph distillation is to learn a small synthetic graph dataset while considering the downstream classification task. There are also several graph-specific approaches designed explicitly for graphs. The eigenbasis matching method proposed in \cite{liu2023graph} aligns the eigenbasis and node features of real and synthetic graphs to enhance the efficiency and generalization of GNN training. However, it may underperform on certain large-scale graphs due to the limited number of eigenvectors available for eigenbasis matching. In \cite{gupta2023mirage}, the MIRAGE approach mines frequently co-occurring computation trees in graphs to distill a smaller, more informative dataset for training GNNs. This approach achieves higher prediction accuracy, significantly improved distillation efficiency, and superior data compression rates. However, specifying the reduction ratio is challenging. Furthermore, the MIRAGE approach assumes that the downstream GNN must be a message-passing network, which may not hold for models such as graph transformers \cite{shirzad2023exphormer}. DosCond \cite{jin2022condensing} assumes discrete graph structures as probabilistic models and performs one-step gradient matching without training the network weights. This method accelerates the condensation process. However, it still relies on gradient-matching similar to traditional methods which might not capture the true distribution of the full graph dataset.

\textbf{Attention Maps.} Attention mechanisms have been extensively utilized in the deep learning community, especially in natural language processing \cite{bahdanau2014neural} and computer vision (CV) \cite{wang2018non,zagoruyko2016paying}. These mechanisms have also been applied to dataset distillation in CV to capture a better representation of the dataset for distillation purposes \cite{sajedi2023datadam, khaki2024atom}. The work by \cite{sajedi2023datadam} demonstrated significant improvements in accuracy through the use of spatial attention across different model layers for matching. However, the application of attention mechanisms in graph condensation remains unexplored. To bridge this gap, we propose structured attention matching in our \texttt{GSTAM} to approximate the distribution of the original graph dataset, reducing the reliance of the distillation on model architecture, and increasing the generality of the learned synthetic graphs.

\subsection{Structural Attention-maps in GNNs}
It has been shown in \cite{zagoruyko2016paying} that activations following convolutional layers in a convolutional neural network highlight the importance of spatial information in the input image. In a similar vein, we argue that this is also true for GNNs and graph inputs. Consider a GNN model with three Graph Convolution (GC) layers. \cref{fig:overall_attention} illustrates the structural attention maps for two different samples from the NCI1 and MUTAG datasets \cite{hu2020open}. These attention maps can be generated using an approach similar to Grad-CAM \cite{selvaraju2017grad}. As seen, the different layers of the GNN focus on different parts of the input graph. For instance, the first layer emphasizes small branches of the molecule, while the third layer focuses more on the overall structure. This trend is also observed in image inputs \cite{zagoruyko2016paying}. In \texttt{GSTAM}, we intend to incorporate such structural information in the distillation process to create a more effective distilled graph dataset.

\section{Methodology}
In this section, we introduce the proposed graph dataset distillation method, \texttt{GSTAM}, which leverages attention maps generated at each layer of a GNN. The core idea behind \texttt{GSTAM} is that each layer of a GNN focuses on distinct structural features of the input graph, analogous to spatial attention mechanisms used in computer vision \cite{sajedi2023datadam, zagoruyko2016paying}. By aligning the structural attention across various layers of the GNN (initial, intermediate, and final layers) trained on both the full and synthetic graph datasets, we can guide the synthetic graph generation process to produce datasets that are more generalized and effective for training GNNs on downstream tasks. The overall procedure of \texttt{GSTAM} is depicted in \cref{fig:attention}

\subsection{Graph Distillation Via Structural Attention Matching}
Suppose we have a graph classification dataset $\mathcal{T}=\{(\mathcal{G}_1, y_1),...,(\mathcal{G}_N, y_N)\}$, where $\mathcal{G}_i$ is the graph sample associated with label $y_i$. Each graph sample $\mathcal{G}_i$ consists of an adjacency matrix $\mathbf{A}_i \in \mathbb{R}^{m_i \times m_i}$ and a node feature matrix $\mathbf{X}_i \in \mathbb{R}^{m_i \times d}$, where $m_i$ denotes the number of nodes in $\mathcal{G}_i$ and $d$ is the feature dimension for each node. Now, consider a synthetic dataset $\mathcal{S}=\{(\mathcal{G}'_1, y_1),...,(\mathcal{G}'_M, y_M)\}$ with $M \ll N$. In the synthetic dataset, each graph $\mathcal{G}'_i$ has an adjacency matrix $\mathbf{A}'_i \in \mathbb{R}^{n \times n}$ and a node feature matrix $\mathbf{X}'_i \in \mathbb{R}^{n \times d}$, where $n$ is set as $n = \frac{1}{N}\sum_{i=1}^{N} m_i$ and the feature dimension $d$ is the same as in the original graphs. The synthetic graphs $\mathcal{G}'_i$ can be initialized randomly or by using a portion of the nodes from the graph samples selected via the K-Center method \cite{farahani2009facility,sener2017active}.

Consider the $c^{\text{th}}$ class in the dataset and a randomly initialized GNN model $GNN_{\boldsymbol{\theta}}(\cdot)$ parameterized by $\boldsymbol{\theta}_l$ for $l=1,...,L$, where $l$ indicates the $l^{\text{th}}$ layer. In most GNN models, $L$ is a small number to prevent the over-smoothing problem caused by deeper layers \cite{keriven2022not}. First, we feed $\mathcal{T}_c$, the subset of $\mathcal{T}$ containing only class $c$ samples, to the GNN model to obtain $GNN_{\boldsymbol{\theta}}(\mathcal{T}_c) = [f^{\mathcal{T}_c}_{\boldsymbol{\theta}_1},..., f^{\mathcal{T}_c}_{\boldsymbol{\theta}_L}]$. Similarly, for the synthetic dataset $\mathcal{S}_c$, we get $GNN_{\boldsymbol{\theta}}(\mathcal{S}_c) = [f^{\mathcal{S}_c}_{\boldsymbol{\theta}_1},..., f^{\mathcal{S}_c}_{\boldsymbol{\theta}_L}]$. Here, $f^{\mathcal{T}_c}_{\boldsymbol{\theta}_l}$ and $f^{\mathcal{S}_c}_{\boldsymbol{\theta}_l}$ are the feature maps of different GNN layers after the activation function, with dimensions $\mathbb{R}^{|\boldsymbol{B}^{\mathcal{T}_c}| \times m_i \times u_l}$ and $\mathbb{R}^{|\boldsymbol{B}^{\mathcal{S}_c}| \times n \times u_l}$, respectively. $\boldsymbol{B}^{\mathcal{T}_c}$ and $\boldsymbol{B}^{\mathcal{S}_c}$ denote the batch of samples for the full and synthetic graph datasets in class $c$, respectively. Additionally, $u_l$ represents the feature dimension in the $l^{\text{th}}$ layer. Without loss of generality, we assume there is no pooling layer in the GNN models, meaning the number of nodes in each graph sample remains the same throughout the layers of the GNN model.
\begin{figure} [t!]
    \centering
    \includegraphics[width=.9\textwidth]{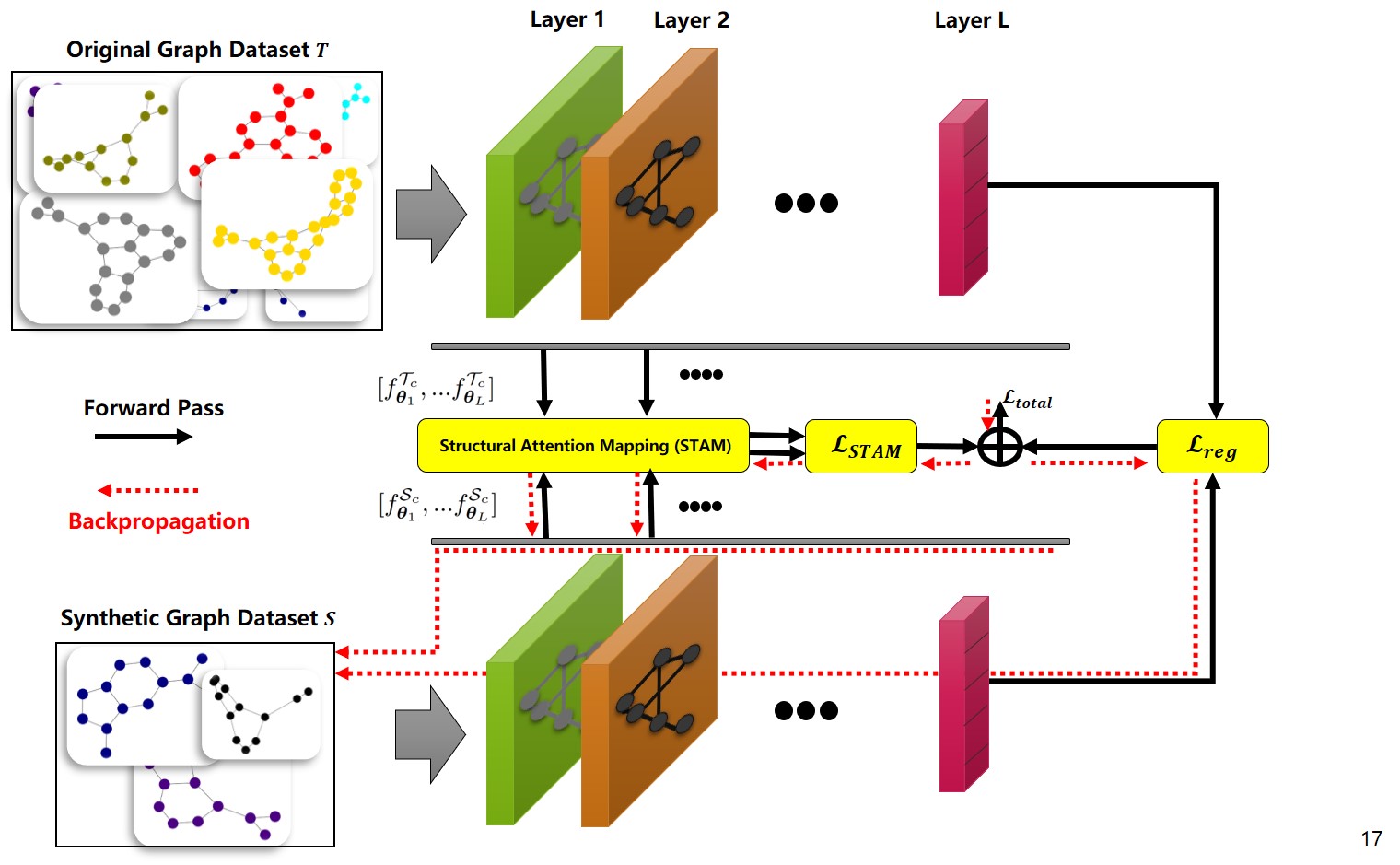}  
    \caption{\textbf{Overveiw of \texttt{GSTAM}:} \texttt{GSTAM} matches the structural attention maps of different layers of a GNN model trained on the full and the synthetic graph dataset, respectively along with the reg loss to account for the final layer of the GNN model.}
    \label{fig:attention}
\end{figure}
\subsubsection{Structural Attention Maps.} We now introduce our \textbf{St}ructural \textbf{A}ttention \textbf{M}ap (\textbf{STAM}) module by defining a feature-based mapping function $A(\cdot)$. This special function creates structural attention maps for the feature maps of different layers. We define $A\left(GNN_{\boldsymbol{\theta}}(\mathcal{T}_c)\right)=[a_{\boldsymbol{\theta}_1}^{\mathcal{T}_c},...,a_{\boldsymbol{\theta}_L}^{\mathcal{T}_c}]$ and $A\left(GNN_{\boldsymbol{\theta}}(\mathcal{S}_c)\right)=[a_{\boldsymbol{\theta}_1}^{\mathcal{S}_c},...,a_{\boldsymbol{\theta}_L}^{\mathcal{S}_c}]$ as the attention maps of the original and synthetic graph datasets transformed with STAM. Conventionally, it was shown in \cite{zagoruyko2016paying, sajedi2023datadam} that a good and flexible choice of the mapping function is a spatial-wise aggregation of the $f^{\mathcal{T}_c}_{\boldsymbol{\theta}_l}$, i.e., $A(f^{\mathcal{T}_c}_{\boldsymbol{\theta}_l}) = a_{\boldsymbol{\theta}_l}^{\mathcal{T}_c} = \sum_{i=1}^{Ch_i} |(f^{\mathcal{T}_c}_{\boldsymbol{\theta}_l})_i|^p$, where $Ch_i$ is the channel number and $p$ controls the attention map. However, in GNNs, the mapping function cannot be defined naively since the dimensionality of $f^{\mathcal{T}_c}_{\boldsymbol{\theta}_l}$ and $f^{\mathcal{S}_c}_{\boldsymbol{\theta}_l}$ are not consistent, i.e., $f^{\mathcal{T}_c}_{\boldsymbol{\theta}_l} \in \mathbb{R}^{|\boldsymbol{B}^{\mathcal{T}_c}| \times m_i \times u_l}$ and $f^{\mathcal{S}_c}_{\boldsymbol{\theta}_l} \in \mathbb{R}^{|\boldsymbol{B}^{\mathcal{S}_c}| \times n \times u_l}$. To alleviate this problem, we propose the following mapping function:
\begin{equation} \label{eq:attention}
    A(f^{\mathcal{T}_c}_{\boldsymbol{\theta}_l}) = \left| \left(f^{\mathcal{T}_c}_{\boldsymbol{\theta}_l}\right)^T \right|^{P} \left|f^{\mathcal{T}_c}_{\boldsymbol{\theta}_l}\right|^{P},
\end{equation}
where $A(f^{\mathcal{T}_c}_{\boldsymbol{\theta}_l}) = a_{\boldsymbol{\theta}_l}^{\mathcal{T}_c} \in \mathbb{R}^{|\boldsymbol{B}^{\mathcal{T}_c}| \times u_l \times u_l}$. A similar mapping function creates $a_{\boldsymbol{\theta}_l}^{\mathcal{S}_c} \in \mathbb{R}^{|\boldsymbol{B}^{\mathcal{S}_c}| \times u_l \times u_l}$. The resultant attention maps highlight the important sub-structures on the graph associated with the neuron, which can be further utilized in the distillation process. This is because the absolute value of hidden neurons can reveal the importance of different parts of the input data, as shown in \cite{zagoruyko2016paying, sajedi2023datadam}. The impact of $p$ will be investigated in Section \cref{sec: ablation}.

\subsubsection{Distillation Losses.} we compare the normalized spatial attention maps of each layer (excluding the final layer) between $\mathcal{T}_c$  and $\mathcal{S}_c$ as follows
\begin{equation}\label{lstam}
\mathcal{L}_{STAM}=\mathbb{E}_{\boldsymbol{\theta} \sim P_{\boldsymbol{\theta}}} \left[ \sum_{c=1}^C \sum_{l=1}^{L-1} \left\| \frac{1}{|\boldsymbol{B}^{\mathcal{T}_c}|} \sum_{i=1}^{|\boldsymbol{B}^{\mathcal{T}_c}|} \frac{(z^{\mathcal{T}_c}_{\boldsymbol{\theta}_l})_i}{\|(z^{\mathcal{T}_c}_{\boldsymbol{\theta}_l})_i\|_2} - \frac{1}{|\boldsymbol{B}^{\mathcal{S}_c}|} \sum_{i=1}^{|\boldsymbol{B}^{\mathcal{S}_c}|} \frac{(z^{\mathcal{S}_c}_{\boldsymbol{\theta}_l})_i}{\|(z^{\mathcal{S}_c}_{\boldsymbol{\theta}_l})_i\|_2}  \right\|^2 \right],
\end{equation}
where $z^{\mathcal{S}_c}_{\boldsymbol{\theta}_l}$ and $z^{\mathcal{T}_c}_{\boldsymbol{\theta}_l}$ denote the vectotized versions of $a_{\boldsymbol{\theta}_l}^{\mathcal{S}_c}$ and $a_{\boldsymbol{\theta}_l}^{\mathcal{T}_c}$ respectively, with the same dimension. In addition the subscript $i$ indicates $(z^{\mathcal{S}_c}_{\boldsymbol{\theta}_l})_i = z^{\mathcal{S}_c}_{\boldsymbol{\theta}_l}(i,:,:)$. Note that normalizing $z_{\boldsymbol{\theta}_l}$ has been shown to be useful in the distillation process \cite{khaki2024atom}. In addition, $P_{\boldsymbol{\theta}}$ represents the distribution of the GNN's parameters.

Although $\mathcal{L}_{STAM}$ effectively captures the distribution of the original graph dataset, it does not account for the last layer of the GNN. Research has consistently shown that the final layer encapsulates abstract and semantic information of the input \cite{saito2018maximum, zhao2023dataset, ma2015hierarchical}. This layer in the case of GNNs is often a linear layer situated after the pooling operation and before the activation function. To address this, we introduce an additional loss based on the last layer of the GNN, comparing the full and synthetic graph datasets using the L2-norm. This loss is formulated as:
\begin{equation} \label{lreg}
\mathcal{L}_{reg} = \mathbb{E}_{\boldsymbol{\theta} \sim P_{\boldsymbol{\theta}}}  \left[ \sum_{c=1}^C \left\| \frac{1}{|\boldsymbol{B}^{\mathcal{T}_c}|} \sum_{i=1}^{|\boldsymbol{B}^{\mathcal{T}_c}|}  (\tilde{f}^{\mathcal{T}_c}_{\boldsymbol{\theta}_L})_i  - \frac{1}{|\boldsymbol{B}^{\mathcal{S}_c}|} \sum_{i=1}^{|\boldsymbol{B}^{\mathcal{S}_c}|}  (\tilde{f}^{\mathcal{S}_c}_{\boldsymbol{\theta}_L})_i \right\|^2 \right],
\end{equation}
where $\tilde{f}^{\mathcal{T}_c}_{\boldsymbol{\theta}_L} \in \mathbb{R}^{|\boldsymbol{B}^{\mathcal{T}_c}|  \times u_L} $ and $\tilde{f}^{\mathcal{S}_c}_{\boldsymbol{\theta}_L} \in \mathbb{R}^{|\boldsymbol{B}^{\mathcal{S}_c}|  \times u_L} $ represent the feature maps of the last layer of the GNN models trained on the full and synthetic graph dataset and the subscript $i$ has the same meaning as explained in $\mathcal{L}_{STAM}$.
\subsubsection{Adjacency Matrix Optimization.}
To update the adjacency matrix, we follow the procedure described by Jin et al. \cite{jin2022condensing}. We assume that the adjacency matrices follow a Bernoulli distribution, defined as $ P_{\boldsymbol{\Omega}_k}(\boldsymbol{A}'_{ij}) = \boldsymbol{A}'_{ij} \sigma(\boldsymbol{\Omega}_{ij}) + (1 - \boldsymbol{A}'_{ij}) \sigma(-\boldsymbol{\Omega}_{ij}) $, where $\boldsymbol{\Omega}_k$ represents the success probability matrix of the Bernoulli distribution for the $k^\textrm{th}$ adjancecy matrix and its values are learned during the optimization process. Moreover, the function $\sigma(\cdot)$ denotes the sigmoid function. It should be noted that in graph classification tasks, the number of nodes in each graph sample is not typically large. This ensures that the time complexity of constructing the adjacency matrix remains manageable and does not adversely affect the overall computational efficiency.
\begin{algorithm}[t!]
    \SetKwInOut{Input}{Input}
    \SetKwInOut{Output}{Output}
    \Input{Full training graph dataset $\mathcal{T} = \{ (\mathcal{G}_i, y_i)\}_{i=1}^{N}$. Initialized synthetic samples for $c$ classes,graph neural network model $GNN_{\boldsymbol{\theta}} (\cdot)$ with parameters $\boldsymbol{\theta}$, Probability distribution over randomly initialized weights $P_{\boldsymbol{\theta}}$, feature and adjacency matrix learning rate $\eta_S$ and $\zeta_S$, Task balance parameter $\lambda$, parameter $p$, Number of training iterations $T$.}

    -Initialize synthetic dataset $\mathcal{S}$
    
     \For{$i = 1, 2, \ldots, T$}{
     - Sample mini-batch pairs $B^\mathcal{T}_k$ and $B^\mathcal{S}_k$ from the real and synthetic sets for each class $c$\\
     - Compute $\mathcal{L}_{STAM}$ and $\mathcal{L}_{reg}$ using \cref{lstam} and \cref{lreg}\\
     - Minimize $\mathcal{L} = \mathcal{L}_{STAM} + \lambda \mathcal{L}_{reg}$ with respect to node features, $\boldsymbol{X}'_i$ and parameters, $\boldsymbol{\Omega}_i$\\
     - Use the $\nabla \mathcal{L}$ to update $\boldsymbol{X}'_i$ and $\boldsymbol{\Omega}_i$ and consequently $\mathcal{S}$

}                 
     \Output{Synthetic dataset $\mathcal{S} = \{(\mathcal{G}'_i, y_i)\}_{i=1}^M$} 
    \caption{\textsc{The Proposed GSTAM}} \label{alg:GSTAM}
\end{algorithm}
Finally, we find a set of synthetic graphs by minimizing the following objective using SGD:
\begin{equation}\label{eq:loss}
    \mathcal{S}^* = \arg\min_{\mathcal{X}',\mathcal{K} } \mathcal{L}_{STAM} + \lambda \mathcal{L}_{reg},
\end{equation}
where $\lambda$ denotes the trade-off parameter, $\mathcal{X}'=[\boldsymbol{X}'_1,...,\boldsymbol{X}'_n]$ is the feature nodes vector, and $\mathcal{K}=[\boldsymbol{\Omega}_1,...,\boldsymbol{\Omega}_M]$ is the parameter vector that determines the adjacency matrices $\{\boldsymbol{A}_i\}_{i=1}^M$ . We investigate the impact of $\lambda$ in Section \cref{sec: ablation}. The summary of the proposed \texttt{GSTAM} is provided in \cref{alg:GSTAM}.

\section{Experiments}

\subsection{Experimental Setup}
\subsubsection{Graph Datasets.}
We evaluate the proposed \texttt{GSTAM} alongside other algorithms on graph classification datasets. In line with previous studies \cite{jin2022condensing, gupta2023mirage}, we select molecular datasets from the Open Graph Benchmark (OGB) (ogbg-molhiv, ogbg-molbbbp) \cite{morris2020tudataset} and TU datasets (DD, MUTAG, NCI1) \cite{hu2020open}. A summary of these datasets is provided in \cref{tab:dataset}. For the DD, MUTAG, and NCI1 datasets, we randomly sample and split the data into $80\%$ for training, $10\%$ for validation, and $10\%$ for testing. For the ogbg-molhiv and ogbg-molbbbp datasets, we use the splits provided by OGB. 

\begin{table} 
    \centering
    \caption{Dataset Descriptions used in the experiments.} \label{tab:dataset}
    \begin{tabular}{cccccc}
    \toprule
        ~~Dataset~~&~~\#Class~~&~~\#Graphs &~~Avg. Node~~&~~Avg. Edges~~&~~Domain~~\\ \hline
         DD  & 2 & 1178 &284.3  & 715.7&Protein\\
         NCI1 & 2 &4110 & 29.9 &32.3 &Molecule\\
         MUTAG & 2 & 187& 18.03 &19.79 &Molecule\\
         ogbg-molhiv & 2 &41,127 & 25.5 & 54.9&Molecule\\
         ogbg-molbbbp & 2 & 2039& 24.1 &26.0 &Molecule\\
         
    \bottomrule
    \end{tabular}
    
    \label{tab:my_label}

\end{table}

\subsubsection{Network Architectures and Implementation Details.}
Similar to the work in \cite{jin2022condensing}, we select a 3 convolutional layer-GNN model with the feature size of $128$ followed by a fully connected layer with $128$ neurons. We followed the same procedure as indicated in \cite{jin2022condensing}. The total number of iterations is set to $T=1000$ where in each iteration a randomly initialized parameter for the GNN model is chosen according to the distribution $P_{\boldsymbol{\theta}}$, i.e., $\boldsymbol{\theta} \sim P_{\boldsymbol{\theta}}$. The feature and adjacency matrix learning rates are set to $\eta_S=0.005$ and $\zeta_S = 0.01$, respectively. Furthermore, we set the task balance parameter $\lambda$ to $0.1$ and $p$ to $2$. The effect of both parameters is studied in \cref{sec: ablation}.
\subsubsection{Baselines.} We compared the proposed \texttt{GSTAM} against three Coreset methods that select graph samples from the full graph dataset, namely \textit{Random}, \textit{Herding} \cite{welling2009herding}, and \textit{K-Center}\cite{farahani2009facility,sener2018active}, and two distillation algorithms that learn the distilled synthetic graphs, such as \textit{DCG} \cite{zhao2021dataset} and \textit{DosCond} \cite{jin2022condensing}. In \textit{Random}, we select the graph samples for a particular class randomly. In \textit{Herding}, we iteratively select samples that are closest to the mean feature representation of the class. In \textit{K-Center}, we choose samples that maximize the minimum distance to the selected set, ensuring diversity. For DCG and DosCond, the same protocol and hyper-parameters are used as described in \cite{jin2022condensing} to have a fair comparison. 

\subsubsection{Evaluation.} The evaluation process involves training on the distilled graphs and testing on the original test set as a graph classification problem. Specifically, we first run the \texttt{GSTAM} algorithm, as detailed in \cref{alg:GSTAM}, to generate the distilled graphs. Next, we train a randomly instantiated GNN model using these distilled graphs with the learning rate of $0.001$ for $500$ epochs (expect for ogbg-molhiv which is $100$ epochs). Finally, we test the trained GNN on the original test dataset. For the DD, MUTAG, and NCI1 datasets, the evaluation metric used is accuracy, while for the others, we use ROC-AUC due to their imbalanced nature. The same procedure is applied to the baselines, where we construct the distilled graphs, train a GNN, and then test it. The distillation process is repeated five times, and the GNN is trained on 10 different randomly instantiated models.

\subsection{Comparison To State-of-the-art Methods}
\begin{table}[bt!]
\caption{Comparison of classification performance against baseline models. The accuracy (in percentages)  is reported for the first three datasets, while ROC-AUC is presented for the remaining datasets. "Full Dataset" refers to the performance using the original graph dataset.}
\label{tab:SOTACOMP}
\resizebox{\textwidth}{!}{%
\begin{tabular}{@{}ccc|ccccccc@{}}
\toprule
Algorithms & \begin{tabular}[c]{@{}c@{}}Graph\\ /class\end{tabular} & Ratio & Random & \begin{tabular}[c]{@{}c@{}}Herding\\ \cite{welling2009herding}\end{tabular} & \begin{tabular}[c]{@{}c@{}}K-Center\\ \cite{sener2018active}\end{tabular} & \begin{tabular}[c]{@{}c@{}}DCG\\ \cite{zhao2021dataset}\end{tabular} & \begin{tabular}[c]{@{}c@{}}DosCond\\ \cite{jin2022condensing}\end{tabular} & GSTAM & Full Dataset \\ \midrule
\multicolumn{1}{c|}{} & 1 & 0.2\% & $0.580\pm 0.067$ & $0.548 \pm 0.034$ & $0.548 \pm 0.034$ & $58.81 \pm 2.90$ & \cellcolor[HTML]{E1F6E1}$70.45 \pm 2.53$ & \multicolumn{1}{c|}{\cellcolor[HTML]{A9FEA8}$70.90\pm 1.64$} & $78.92 \pm 0.64$ \\
\multicolumn{1}{c|}{} & 10 & 2.1\% & $64.69 \pm 2.55$ & $69.79 \pm 2.30$ & $63.46 \pm 2.38$ & $61.84 \pm 1.44$ & \cellcolor[HTML]{A9FEA8}$73.53\pm 1.13$ & \multicolumn{1}{c|}{\cellcolor[HTML]{E1F6E1}$72.11\pm 2.34$} &  \\
\multicolumn{1}{c|}{\multirow{-3}{*}{\begin{tabular}[c]{@{}c@{}}DD \cite{hu2020open}\\ (Accuracy)\end{tabular}}} & 50 & 10.6\% & $67.29 \pm 1.53$ & \cellcolor[HTML]{E1F6E1}$73.95 \pm 1.70$ & $67.41 \pm 0.92$ & $61.27 \pm 1.01$ & \cellcolor[HTML]{A9FEA8}$77.04\pm 1.86$ & \multicolumn{1}{c|}{$72.59\pm 2.88$} &  \\ \midrule
\multicolumn{1}{c|}{} & 1 & 1.3\% & $67.47 \pm9.74$ & $70.84 \pm 7.71$ & $70.84 \pm 7.71$ & $75.00 \pm 8.16$ & \cellcolor[HTML]{E1F6E1}$82.31 \pm 1.21$ & \multicolumn{1}{c|}{\cellcolor[HTML]{A9FEA8}$89.05 \pm 4.29$} &  \\
\multicolumn{1}{c|}{} & 10 & 13.3\% & $77.89 \pm 7.55$ & $80.42 \pm 1.89$ & $81.00 \pm 2.51$ & $82.66 \pm 0.68$ & \cellcolor[HTML]{E1F6E1}$82.56 \pm 2.01$ & \multicolumn{1}{c|}{\cellcolor[HTML]{A9FEA8}\textbf{$89.47\pm 6.03$}} &  \\
\multicolumn{1}{c|}{\multirow{-3}{*}{\begin{tabular}[c]{@{}c@{}}MUTAG \cite{hu2020open} \\ (Accuracy)\end{tabular}}} & 20 & 26.7\% & $78.21 \pm 5.31$ & $80.00 \pm 1.10$ & $82.97 \pm 4.91$ & $82.89 \pm 1.03$ & \cellcolor[HTML]{E1F6E1}$83.21\pm 2.33$ & \multicolumn{1}{c|}{\cellcolor[HTML]{A9FEA8}$84.33\pm 3.93$} & \multirow{-3}{*}{$88.63 \pm 1.44$} \\ \midrule
\multicolumn{1}{c|}{} & 1 & 0.1\% & $51.27 \pm 1.22$ & $53.98 \pm 0.67$ & $53.98 \pm 0.67$ & $51.14 \pm 1.08$ & \cellcolor[HTML]{E1F6E1}$56.60 \pm 0.48$ & \multicolumn{1}{c|}{\cellcolor[HTML]{A9FEA8}$57.05\pm 0.81$} &  \\
\multicolumn{1}{c|}{} & 10 & 0.6\% & $54.33 \pm 3.14$ & $57.11 \pm 0.56$ & $53.21 \pm 1.44$ & $51.86 \pm 0.81$ & \cellcolor[HTML]{A9FEA8}$58.02 \pm 1.05$ & \multicolumn{1}{c|}{\cellcolor[HTML]{E1F6E1}$57.33 \pm 0.41$} &  \\
\multicolumn{1}{c|}{\multirow{-3}{*}{\begin{tabular}[c]{@{}c@{}}NCI1 \cite{hu2020open}\\ (Accuracy)\end{tabular}}} & 50 & 3.0\% & $58.51 \pm 1.73$ & $58.94 \pm 0.83$ & $56.58 \pm 3.08$ & $52.17 \pm 1.90$ & \cellcolor[HTML]{E1F6E1}$60.07 \pm 1.58$ & \multicolumn{1}{c|}{\cellcolor[HTML]{A9FEA8}$61.39 \pm 1.29$} & \multirow{-3}{*}{$71.70 \pm 0.20$} \\ \midrule
\multicolumn{1}{c|}{} & 1 & 0.01\% & $0.719 \pm 0.009$ & $0.721 \pm 0.002$ & $0.721 \pm 0.002$ & $0.718 \pm 0.013$ & \cellcolor[HTML]{E1F6E1}$0.726 \pm 0.003$ & \multicolumn{1}{c|}{\cellcolor[HTML]{A9FEA8}$0.732 \pm 0.002$} &  \\
\multicolumn{1}{c|}{} & 10 & 0.06\% & $0.720 \pm 0.011$ & $0.725 \pm 0.006$ & $0.713 \pm 0.009$ & $0.728 \pm 0.002$ & \cellcolor[HTML]{E1F6E1}$0.728 \pm 0.005$ & \multicolumn{1}{c|}{\cellcolor[HTML]{A9FEA8}$0.734 \pm 0.010$} &  \\
\multicolumn{1}{c|}{\multirow{-3}{*}{\begin{tabular}[c]{@{}c@{}}ogbg-molhiv \cite{morris2020tudataset}\\ (ROC-AUC)\end{tabular}}} & 50 & 0.3\% & $0.721 \pm 0.041$ & $0.725 \pm 0.003$ & $0.725 \pm 0.006$ & $0.726 \pm 0.010$ & \cellcolor[HTML]{E1F6E1}$0.731 \pm 0.004$ & \multicolumn{1}{c|}{\cellcolor[HTML]{A9FEA8}$0.737\pm 0.010$} & \multirow{-3}{*}{\textbf{$0.757 \pm 0.007$}} \\ \midrule
\multicolumn{1}{c|}{} & 1 & 0.1\% & $0.519 \pm 0.016$ & $0.546 \pm 0.019$ & $0.546 \pm 0.019$ & $0.559 \pm 0.044$ & \cellcolor[HTML]{E1F6E1}$0.581\pm 0.005$ & \multicolumn{1}{c|}{\cellcolor[HTML]{A9FEA8}$0.602\pm 0.006$} &  \\
\multicolumn{1}{c|}{} & 10 & 1.2\% & $0.586 \pm 0.040$ & \cellcolor[HTML]{E1F6E1}$0.606\pm 0.019$ & $0.530\pm 0.039$ & $0.568\pm 0.032$ & \cellcolor[HTML]{E1F6E1}$0.606\pm 0.008$ & \multicolumn{1}{c|}{\cellcolor[HTML]{A9FEA8}$0.642\pm 0.003$} &  \\
\multicolumn{1}{c|}{\multirow{-3}{*}{\begin{tabular}[c]{@{}c@{}}ogbg-molbbbp \cite{morris2020tudataset}\\ (ROC-AUC)\end{tabular}}} & 50 & 6.1\% & $0.606\pm 0.020$ & $0.617\pm 0.003$ & $0.576\pm 0.019$ & $0.579\pm 0.032$ & \cellcolor[HTML]{E1F6E1}$0.620\pm 0.007$ & \multicolumn{1}{c|}{\cellcolor[HTML]{A9FEA8}$0.644\pm 0.005$} & \multirow{-3}{*}{$0.646\pm 0.004$} \\ \bottomrule
\end{tabular}%
}
\end{table}
\subsubsection{Performance Comparison.}
We compared the proposed \texttt{GSTAM} with various coreset and distillation approaches, as shown in \cref{tab:SOTACOMP}. \texttt{GSTAM} significantly outperforms most coreset algorithms. This superior performance is attributed to \texttt{GSTAM}'s ability to learn distilled graphs specifically optimized for downstream graph classification tasks. Additionally, \texttt{GSTAM} achieves higher classification performance compared to SOTA graph distillation algorithms, such as DosCond, across almost all datasets. This improvement is primarily because \texttt{GSTAM} incorporates the attention mechanisms of GNN layers during the distillation process, unlike DosCond.

A detailed analysis reveals that \texttt{GSTAM} excels, particularly with small ratios which makes it a good candidate for creating small-sized synthetic graph datasets. For the MUTAG dataset, \texttt{GSTAM} achieves a lossless performance against the full dataset and $6.9\%$ higher accuracy than DosCond. In addition, our proposed algorithm performs nearly as well as the full dataset on the ogbg-molbbbp dataset. We hypothesize that \texttt{GSTAM}'s lossless performance compared to the full dataset is due to the distillation process reducing the impact of outliers present in the original dataset.   
\begin{table}[tb!]
\centering
\caption{The running time comparison (in minutes). } \label{tab:running_time}
\resizebox{\textwidth}{!}{%
\begin{tabular}{c|ccc|cccc}
\toprule
\multirow{2}{*}{Grph/class} & \multicolumn{3}{c|}{DD} & \multicolumn{3}{c}{ogbg-molhiv} \\
\cline{2-7}
           &~~Herding~~ &~~ DosCond~~ & ~~{GSTAM}~~ & ~~Herding ~~& DosCond ~~&~~ {GSTAM}~~ \\
           \cline{2-7}
1          & 8.6m & 8.2m & 8.5m & 72.1m & 11m & 11.8m \\
10         & 8.6m & 8.2m & 8.5m & 72.1m & 11m & 11.8m \\

50         & 8.6m & 8.3m & 8.6m & 72.1m & 11.2m & 11.9m \\
\bottomrule
\end{tabular}%
}
\end{table}
\subsubsection{Running Time.}
We assess the computational efficiency of our method by comparing it with DosCond and a coreset approach, Herding. Herding is known for its relatively lower time consumption compared to other coreset methods and its generally superior performance to other baseline methods, making it a suitable benchmark. \cref{tab:running_time} presents the running times on different datasets. As observed, \texttt{GSTAM} has a comparable time complexity to DosCond and is much faster than coreset algorithms like Herding. Note that both DosCond and \texttt{GSTAM} were evaluated using the same number of iterations for a fair comparison. Therefore, \texttt{GSTAM} demonstrates efficiency on par with DosCond in terms of running time.
\begin{figure}[tb!]
    \centering
    \subfloat[ogbg-molbbbp]{
        \includegraphics[width=0.45\textwidth]{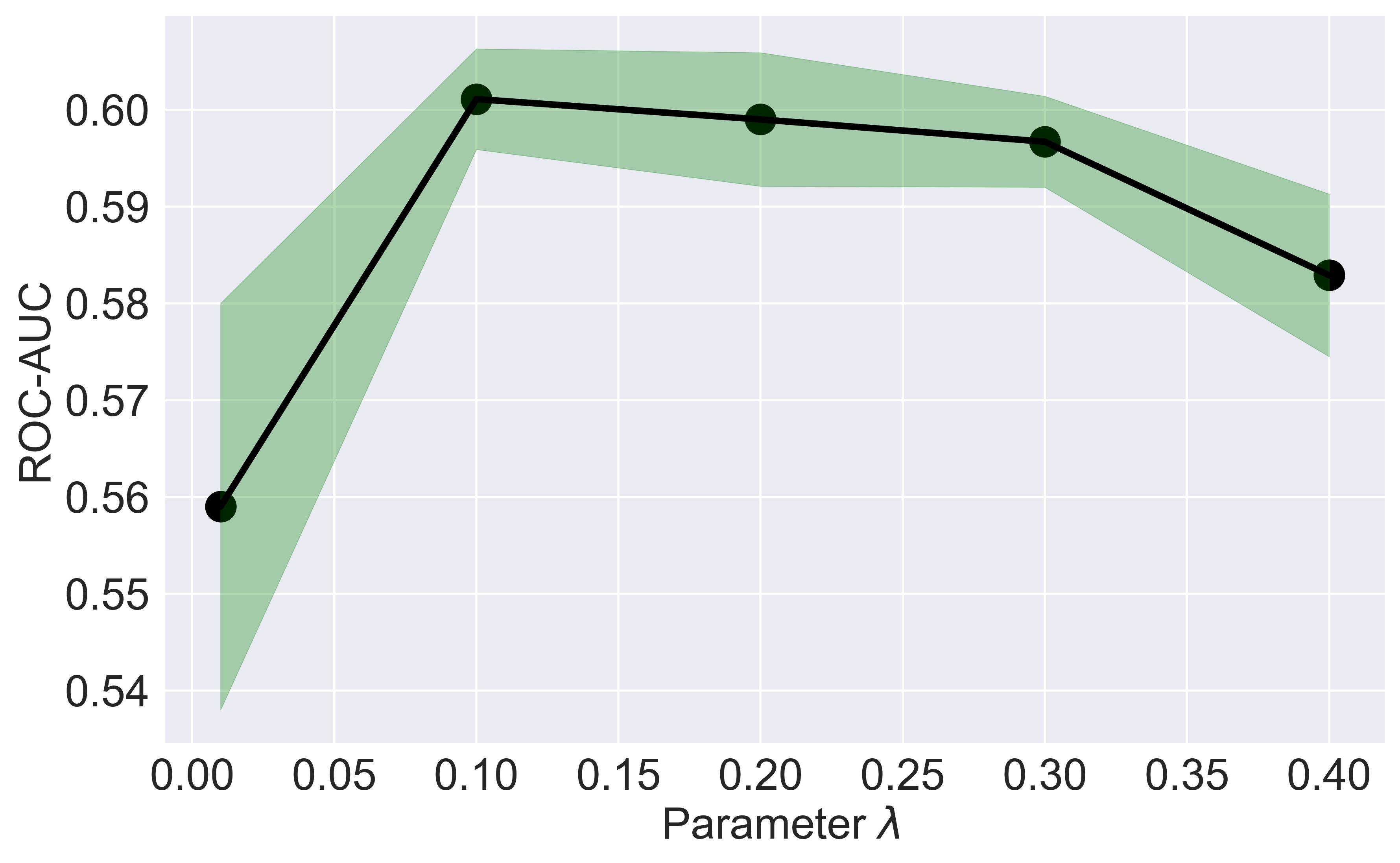}
        \label{fig:subfig1}
    }
    \hfill
    \subfloat[ogbg-molhiv]{
        \includegraphics[width=0.45\textwidth]{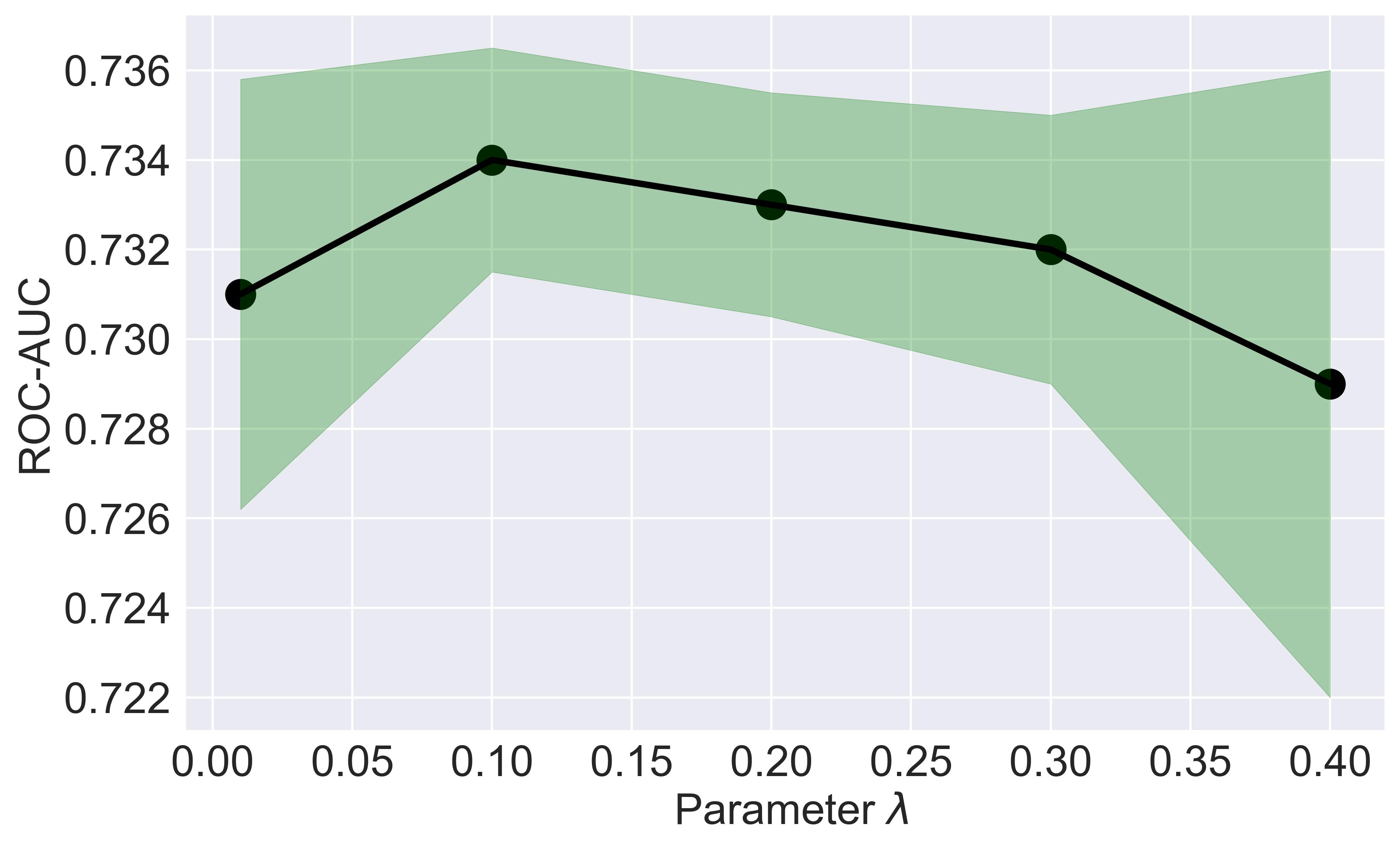}
        \label{fig:subfig2}
    } \\
    \subfloat[ogbg-molbbbp]{
        \includegraphics[width=0.45\textwidth]{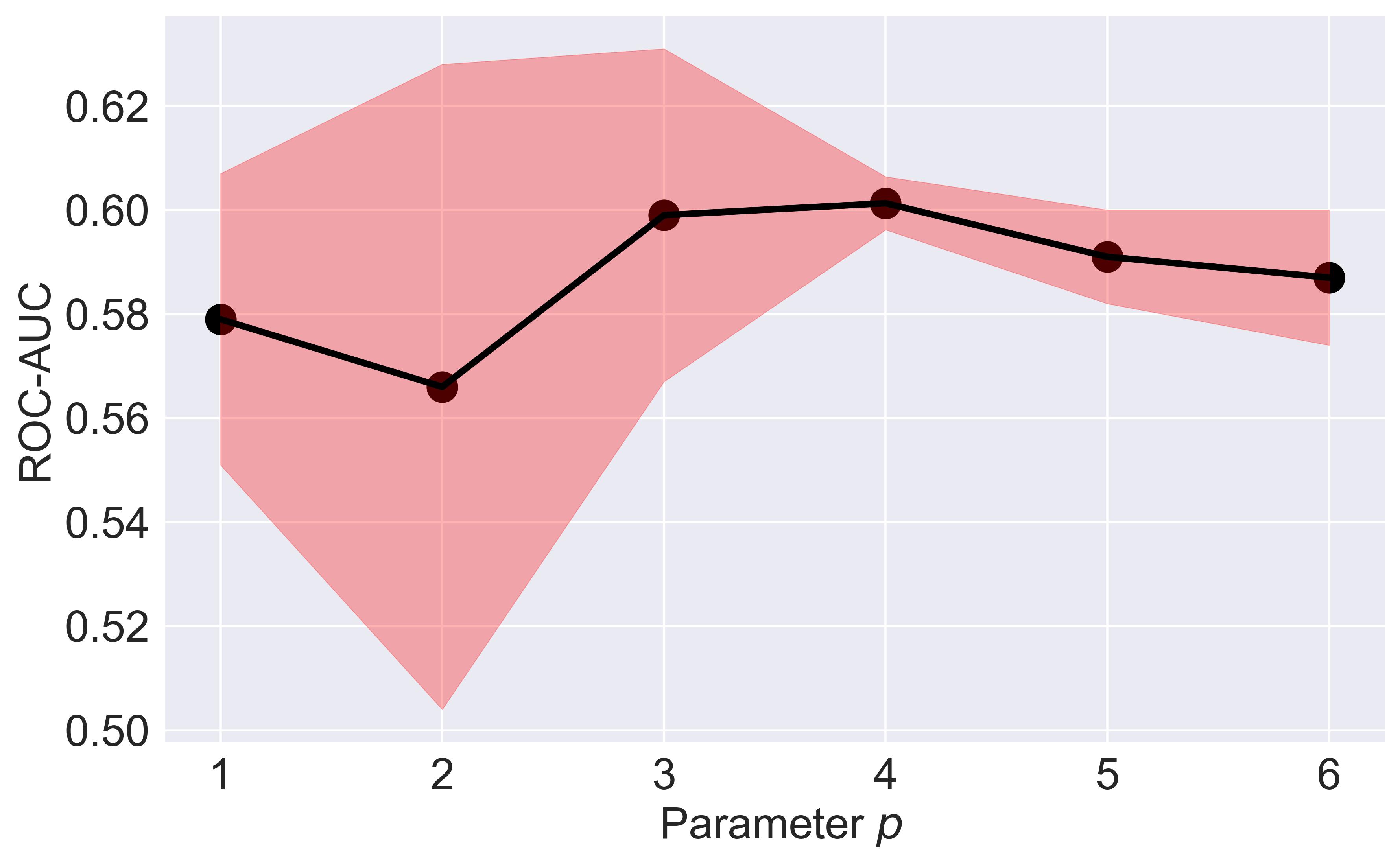}
        \label{fig:subfig3}
    }
    \hfill
    \subfloat[ogbg-molhiv]{
        \includegraphics[width=0.45\textwidth]{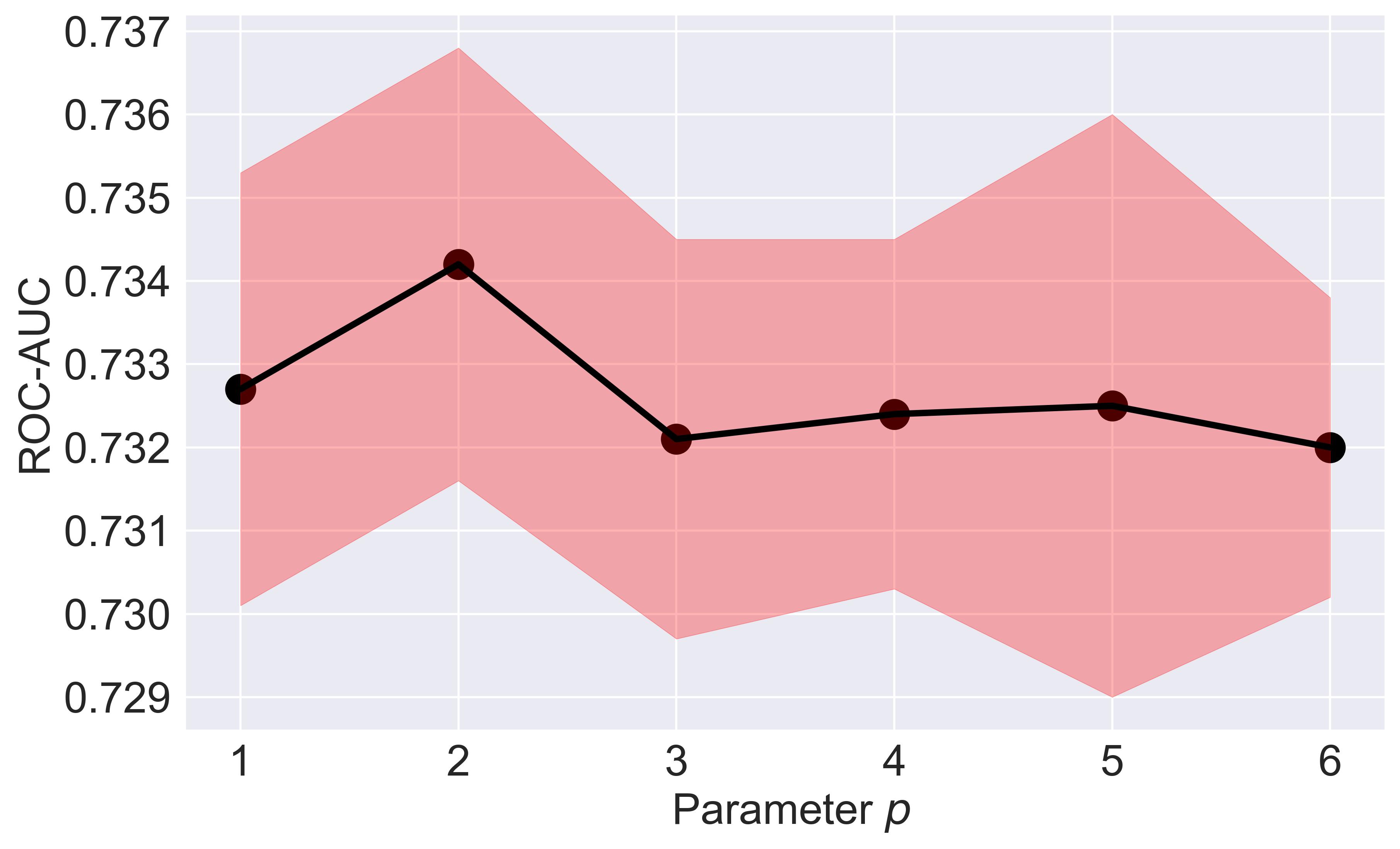}
        \label{fig:subfig4}
    }
    \caption{The effect of different parameters on the ogbg-molbbbp and ogbg-molhiv datasets; (a,b) illustrate the impact of the trade-off parameter $\lambda$ used in \cref{eq:loss}, while (c,d) demonstrate the effect of the parameter $p$ used in \cref{eq:attention}.}
    \label{fig:overall}
\end{figure}
\subsection{Ablation Studies} \label{sec: ablation}
\subsubsection{Exploring the effect of $\lambda$.} The parameter $\lambda$ determines the importance of matching the distribution of the last layer. \cref{fig:subfig1} and \cref{fig:subfig2} illustrate the impact of $\lambda$ on the graph distillation performance for the ogbg-molbbbp and ogbg-molhiv datasets, respectively. When $\lambda$ is small, indicating that $\mathcal{L}_{reg}$ is not significantly considered, the performance is low. Conversely, as $\lambda$ increases, we observe the importance of $\mathcal{L}_{STAM}$ in enhancing performance. However, when $\lambda$ becomes too large, the performance decreases. This suggests that while the structural attention matching loss $\mathcal{L}_{STAM}$ significantly contributes to performance improvement, a small but optimal amount of the regularization term $\mathcal{L}_{reg}$ is necessary to boost performance effectively.
\subsubsection{Exploring the effect of $p$.} The parameter $p$ in \cref{eq:attention} influences the model's ability to catch the attention patterns similar to the ones used in \cite{zagoruyko2016paying, khaki2024atom}. \cref{fig:subfig3} and \cref{fig:subfig4} show the impact of varying $p$ on the graph distillation performance for the ogbg-molbbbp and ogbg-molhiv datasets, respectively. We can see that a certain value for $p$ cannot be determined and the performance is fairly consistent over different values. However, in our experiments, we heuristically set $p=2$ since the distillation overall works better for all datasets.

\subsubsection{Cross-Architecture Analysis.}
To assess the generalizability of the learned graphs, we trained synthetic data on one architecture and tested it using different model architectures. \cref{tab:cross_arch} presents the cross-architecture testing performance. Specifically, we trained the synthetics using a GNN model with two (GCN-2C) and four (GCN-4C) graph convolutional layers on the MUTAG dataset, with one graph per class. The architectures we select for the testing phase are GCN-2C, GCN-3C, a three-layer Graph Isomorphism Network (GIN) \cite{xu2018powerful}, and a three-layer Message Passing Neural Network (MPNN) \cite{gilmer2017neural}.

Although the accuracy of cross-architecture performance decreases for both \texttt{GSTAM} and DosCond, \texttt{GSTAM} still consistently outperforms DosCond across nearly all models. This is attributed to \texttt{GSTAM}'s ability to incorporate the focus of the GNN layers into the synthetic graphs, leading to a better representation of the datasets in the distilled graphs. Consequently, \texttt{GSTAM} demonstrates great generalizability compared to approaches such as gradient matching in DosCond.  

\begin{table}[bt!]
\caption{Cross-architecture testing performance (\%) on MUTAG dataset with 1 graph per class. The synthetic set is trained on one architecture and evaluated on another.}
\label{tab:cross_arch}
\resizebox{\textwidth}{!}{%
\begin{tabular}{@{}c|c|cccc@{}}
\toprule
 & GNN Models & ~~~~~ GCN-2C ~~~~~ & ~~~~~ GCN-3C ~~~~~ & ~~~~~ GIN \cite{xu2018powerful} ~~~~~ & ~~~~~ MPNN \cite{gilmer2017neural} ~~~~~ \\ \midrule
\multirow{2}{*}{DosCond} & GCN-3C & $82.45 \pm 6.21$ & $82.31 \pm 1.21$ & \textbf{85.96 $\pm$ 4.96} & $77.19 \pm 8.90$ \\
 & GCN-4C & $85.96 \pm 8.94$ & $82.45 \pm 6.56$ & $78.94 \pm 3.06$ & $84.21 \pm 2.53$ \\ \midrule
\multirow{2}{*}{GSTAM} & GCN-3C & \textbf{84.21 $\pm$ 7.23} & \textbf{89.05 $\pm$ 4.29} & $82.45 \pm 4.96$ & \textbf{84.11 $\pm$ 8.68} \\
 & GCN-4C & \textbf{88.47 $\pm$ 7.44} & \textbf{87.96 $\pm$ 2.23} & \textbf{80.70 $\pm$ 6.62} & \textbf{87.71 $\pm$ 2.48} \\ \bottomrule
\end{tabular}%
}
\end{table}
\section{Conclusion}
In this paper, we introduced Graph Distillation with Structural Attention Matching (\texttt{GSTAM}), a novel method for condensing graph classification datasets. \texttt{GSTAM} leverages the attention maps of Graph Neural Networks (GNNs) to distill structural information from the original dataset into synthetic graphs. Our comprehensive experiments show that \texttt{GSTAM} not only excels in accuracy, particularly with small condensation ratios compared to the baselines but also matches or surpasses full dataset performance by mitigating outlier impacts in some cases. Further assessment of cross-architecture analysis demonstrates that \texttt{GSTAM} consistently outperforms existing methods like DosCond across various model architectures. These results underscore \texttt{GSTAM}'s potential for enhancing graph classification tasks, making it a valuable tool for performing graph distillation in practical applications. Future work involves applying \texttt{GSTAM} on more complex graph classification datasets as well as node classification tasks. 
\bibliographystyle{unsrt}  
\bibliography{main}

\begin{thebibliography}{10}

\bibitem{hu2020open}
Weihua Hu, Matthias Fey, Marinka Zitnik, Yuxiao Dong, Hongyu Ren, Bowen Liu, Michele Catasta, and Jure Leskovec.
\newblock Open graph benchmark: Datasets for machine learning on graphs.
\newblock {\em Advances in neural information processing systems}, 33:22118--22133, 2020.

\bibitem{ching2015one}
Avery Ching, Sergey Edunov, Maja Kabiljo, Dionysios Logothetis, and Sambavi Muthukrishnan.
\newblock One trillion edges: Graph processing at facebook-scale.
\newblock {\em Proceedings of the VLDB Endowment}, 8(12):1804--1815, 2015.

\bibitem{ying2018graph}
Rex Ying, Ruining He, Kaifeng Chen, Pong Eksombatchai, William~L Hamilton, and Jure Leskovec.
\newblock Graph convolutional neural networks for web-scale recommender systems.
\newblock In {\em Proceedings of the 24th ACM SIGKDD international conference on knowledge discovery \& data mining}, pages 974--983, 2018.

\bibitem{zhang2022cglb}
Xikun Zhang, Dongjin Song, and Dacheng Tao.
\newblock Cglb: Benchmark tasks for continual graph learning.
\newblock {\em Advances in Neural Information Processing Systems}, 35:13006--13021, 2022.

\bibitem{gao2021graph}
Yang Gao, Hong Yang, Peng Zhang, Chuan Zhou, and Yue Hu.
\newblock Graph neural architecture search.
\newblock In {\em International joint conference on artificial intelligence}. International Joint Conference on Artificial Intelligence, 2021.

\bibitem{zhang2023dynamic}
Zeyang Zhang, Ziwei Zhang, Xin Wang, Yijian Qin, Zhou Qin, and Wenwu Zhu.
\newblock Dynamic heterogeneous graph attention neural architecture search.
\newblock In {\em Proceedings of the AAAI Conference on Artificial Intelligence}, volume~37, pages 11307--11315, 2023.

\bibitem{xu2023not}
Peng Xu, Lin Zhang, Xuanzhou Liu, Jiaqi Sun, Yue Zhao, Haiqin Yang, and Bei Yu.
\newblock Do not train it: A linear neural architecture search of graph neural networks.
\newblock In {\em International Conference on Machine Learning}, pages 38826--38847. PMLR, 2023.

\bibitem{jing2021amalgamating}
Yongcheng Jing, Yiding Yang, Xinchao Wang, Mingli Song, and Dacheng Tao.
\newblock Amalgamating knowledge from heterogeneous graph neural networks.
\newblock In {\em Proceedings of the IEEE/CVF conference on computer vision and pattern recognition}, pages 15709--15718, 2021.

\bibitem{jin2022condensing}
Wei Jin, Xianfeng Tang, Haoming Jiang, Zheng Li, Danqing Zhang, Jiliang Tang, and Bing Yin.
\newblock Condensing graphs via one-step gradient matching.
\newblock In {\em Proceedings of the 28th ACM SIGKDD Conference on Knowledge Discovery and Data Mining}, pages 720--730, 2022.

\bibitem{jin2021graph}
Wei Jin, Lingxiao Zhao, Shichang Zhang, Yozen Liu, Jiliang Tang, and Neil Shah.
\newblock Graph condensation for graph neural networks.
\newblock {\em arXiv preprint arXiv:2110.07580}, 2021.

\bibitem{zheng2024structure}
Xin Zheng, Miao Zhang, Chunyang Chen, Quoc Viet~Hung Nguyen, Xingquan Zhu, and Shirui Pan.
\newblock Structure-free graph condensation: From large-scale graphs to condensed graph-free data.
\newblock {\em Advances in Neural Information Processing Systems}, 36, 2024.

\bibitem{zhang2024navigating}
Yuchen Zhang, Tianle Zhang, Kai Wang, Ziyao Guo, Yuxuan Liang, Xavier Bresson, Wei Jin, and Yang You.
\newblock Navigating complexity: Toward lossless graph condensation via expanding window matching.
\newblock {\em arXiv preprint arXiv:2402.05011}, 2024.

\bibitem{liu2022graph}
Mengyang Liu, Shanchuan Li, Xinshi Chen, and Le~Song.
\newblock Graph condensation via receptive field distribution matching.
\newblock {\em arXiv preprint arXiv:2206.13697}, 2022.

\bibitem{zagoruyko2016paying}
Sergey Zagoruyko and Nikos Komodakis.
\newblock Paying more attention to attention: Improving the performance of convolutional neural networks via attention transfer.
\newblock {\em arXiv preprint arXiv:1612.03928}, 2016.

\bibitem{sajedi2023datadam}
Ahmad Sajedi, Samir Khaki, Ehsan Amjadian, Lucy~Z Liu, Yuri~A Lawryshyn, and Konstantinos~N Plataniotis.
\newblock Datadam: Efficient dataset distillation with attention matching.
\newblock In {\em Proceedings of the IEEE/CVF International Conference on Computer Vision}, pages 17097--17107, 2023.

\bibitem{khaki2024atom}
Samir Khaki, Ahmad Sajedi, Kai Wang, Lucy~Z Liu, Yuri~A Lawryshyn, and Konstantinos~N Plataniotis.
\newblock Atom: Attention mixer for efficient dataset distillation.
\newblock In {\em Proceedings of the IEEE/CVF Conference on Computer Vision and Pattern Recognition}, pages 7692--7702, 2024.

\bibitem{agarwal2020contextual}
Sharat Agarwal, Himanshu Arora, Saket Anand, and Chetan Arora.
\newblock Contextual diversity for active learning.
\newblock In {\em Computer Vision--ECCV 2020: 16th European Conference, Glasgow, UK, August 23--28, 2020, Proceedings, Part XVI 16}, pages 137--153. Springer, 2020.

\bibitem{sener2018active}
Ozan Sener and Silvio Savarese.
\newblock Active learning for convolutional neural networks: A core-set approach.
\newblock In {\em International Conference on Learning Representations}, 2018.

\bibitem{welling2009herding}
Max Welling.
\newblock Herding dynamical weights to learn.
\newblock In {\em Proceedings of the 26th annual international conference on machine learning}, pages 1121--1128, 2009.

\bibitem{farahani2009facility}
Reza~Zanjirani Farahani and Masoud Hekmatfar.
\newblock {\em Facility location: concepts, models, algorithms and case studies}.
\newblock Springer Science \& Business Media, 2009.

\bibitem{liu2023graph}
Yang Liu, Deyu Bo, and Chuan Shi.
\newblock Graph condensation via eigenbasis matching.
\newblock {\em arXiv preprint arXiv:2310.09202}, 2023.

\bibitem{gupta2023mirage}
Mridul Gupta, Sahil Manchanda, Sayan Ranu, and Hariprasad Kodamana.
\newblock Mirage: Model-agnostic graph distillation for graph classification.
\newblock {\em arXiv preprint arXiv:2310.09486}, 2023.

\bibitem{shirzad2023exphormer}
Hamed Shirzad, Ameya Velingker, Balaji Venkatachalam, Danica~J Sutherland, and Ali~Kemal Sinop.
\newblock Exphormer: Sparse transformers for graphs.
\newblock In {\em International Conference on Machine Learning}, pages 31613--31632. PMLR, 2023.

\bibitem{bahdanau2014neural}
Dzmitry Bahdanau, Kyunghyun Cho, and Yoshua Bengio.
\newblock Neural machine translation by jointly learning to align and translate.
\newblock {\em arXiv preprint arXiv:1409.0473}, 2014.

\bibitem{wang2018non}
Xiaolong Wang, Ross Girshick, Abhinav Gupta, and Kaiming He.
\newblock Non-local neural networks.
\newblock In {\em Proceedings of the IEEE conference on computer vision and pattern recognition}, pages 7794--7803, 2018.

\bibitem{selvaraju2017grad}
Ramprasaath~R Selvaraju, Michael Cogswell, Abhishek Das, Ramakrishna Vedantam, Devi Parikh, and Dhruv Batra.
\newblock Grad-cam: Visual explanations from deep networks via gradient-based localization.
\newblock In {\em Proceedings of the IEEE international conference on computer vision}, pages 618--626, 2017.

\bibitem{sener2017active}
Ozan Sener and Silvio Savarese.
\newblock Active learning for convolutional neural networks: A core-set approach.
\newblock {\em arXiv preprint arXiv:1708.00489}, 2017.

\bibitem{keriven2022not}
Nicolas Keriven.
\newblock Not too little, not too much: a theoretical analysis of graph (over) smoothing.
\newblock {\em Advances in Neural Information Processing Systems}, 35:2268--2281, 2022.

\bibitem{saito2018maximum}
Kuniaki Saito, Kohei Watanabe, Yoshitaka Ushiku, and Tatsuya Harada.
\newblock Maximum classifier discrepancy for unsupervised domain adaptation.
\newblock In {\em Proceedings of the IEEE conference on computer vision and pattern recognition}, pages 3723--3732, 2018.

\bibitem{zhao2023dataset}
Bo~Zhao and Hakan Bilen.
\newblock Dataset condensation with distribution matching.
\newblock In {\em Proceedings of the IEEE/CVF Winter Conference on Applications of Computer Vision}, pages 6514--6523, 2023.

\bibitem{ma2015hierarchical}
Chao Ma, Jia-Bin Huang, Xiaokang Yang, and Ming-Hsuan Yang.
\newblock Hierarchical convolutional features for visual tracking.
\newblock In {\em Proceedings of the IEEE international conference on computer vision}, pages 3074--3082, 2015.

\bibitem{morris2020tudataset}
Christopher Morris, Nils~M Kriege, Franka Bause, Kristian Kersting, Petra Mutzel, and Marion Neumann.
\newblock Tudataset: A collection of benchmark datasets for learning with graphs.
\newblock {\em arXiv preprint arXiv:2007.08663}, 2020.

\bibitem{zhao2021dataset}
Bo~Zhao, Konda~Reddy Mopuri, and Hakan Bilen.
\newblock Dataset condensation with gradient matching.
\newblock In {\em Ninth International Conference on Learning Representations 2021}, 2021.

\bibitem{xu2018powerful}
Keyulu Xu, Weihua Hu, Jure Leskovec, and Stefanie Jegelka.
\newblock How powerful are graph neural networks?
\newblock In {\em International Conference on Learning Representations}, 2018.

\bibitem{gilmer2017neural}
Justin Gilmer, Samuel~S Schoenholz, Patrick~F Riley, Oriol Vinyals, and George~E Dahl.
\newblock Neural message passing for quantum chemistry.
\newblock In {\em International conference on machine learning}, pages 1263--1272. PMLR, 2017.

\end{thebibliography}

\end{document}